\documentclass[runningheads]{llncs}
\usepackage{graphicx}
\usepackage{amsmath,amssymb} %

\usepackage{color}
\usepackage{epsfig}
\usepackage{graphicx}
\usepackage{subcaption}
\usepackage{enumitem}
\usepackage{multirow}
\usepackage{graphicx}
\usepackage[width=122mm,left=12mm,paperwidth=146mm,height=193mm,top=12mm,paperheight=217mm]{geometry}

\usepackage[font=footnotesize]{caption}
\DeclareCaptionFormat{myformat}{#1#2#3\hrulefill}
\begin{document}

\newcommand{\etal}{\textit{et al}. }
\newcommand{\ie}{\textit{i}.\textit{e}., }
\newcommand{\eg}{\textit{e}.\textit{g}. }

\pagestyle{headings}
\mainmatter
\def\ECCV18SubNumber{2980}  %

\title{Learning to Predict Crisp Boundaries} %

\titlerunning{Learning to predict crisp boundaries}

\authorrunning{Deng et al.}

\author{Ruoxi Deng$^{1,\star}$, Chunhua Shen$^{2}$, Shengjun Liu$^{1}$, Huibing Wang$^{3,\star}$, Xinru Liu$^{1}$}

\institute{$^{1}$Central South University, China;  $^{2}$The University of Adelaide, Australia;  $^{3}$Dalian  University of Technology, China\thanks{Part of this work was done
when R. Deng and H. Wang were visiting The University of Adelaide. Appearing in Proc.\ European Conference on Computer Vision (ECCV), 2018.
 }}

\maketitle

\begin{abstract}

    Recent methods for boundary or edge detection built on Deep Convolutional Neural Networks (CNNs) 
typically suffer from the issue of predicted edges being thick and  need post-processing to obtain crisp boundaries. 
Highly imbalanced categories of boundary versus background in training data is one of main reasons for the above problem.
In this work, the aim is to make CNNs produce sharp boundaries without post-processing. 
We introduce a novel loss for boundary detection, which is very effective for classifying imbalanced data 
and allows CNNs to produce crisp boundaries. 
Moreover, we propose an end-to-end network which adopts the bottom-up/top-down architecture to tackle the task. 
The proposed network effectively leverages hierarchical features and produces pixel-accurate boundary mask, 
which is critical to reconstruct the edge map. 
Our experiments illustrate that directly making crisp prediction not only promotes the visual results of CNNs, 
but also achieves better results against the state-of-the-art 
on the BSDS500 dataset (ODS F-score of .815) and the NYU Depth dataset (ODS F-score of .762).

\end{abstract}

\clearpage 

\section{Introduction}

Edge detection is a long-standing task in computer vision \cite{marr1980theory,gonzalezdigital}. 
In early years, the objective is defined as to find sudden changes of discontinuities in intensity images \cite{torre1986edge}. 
Nowadays, it is  expected to localize semantically meaningful objects’ boundaries, 
which play a fundamental and significant role in many computer vision tasks such as image segmentation \cite{senthilkumaran2009edge,arbelaez2011contour,chen2016semantic,bertasius2016semantic} and optical flow \cite{ren2008local,revaud2015epicflow}. 
In the past few years, deep convolutional neural networks (CNNs) have dominated the research on  edge detection. 
CNN based methods, such as DeepEdge \cite{bertasius2015deepedge}, DeepContour \cite{shen2015deepcontour}, HED \cite{xie2015holistically} and RCF \cite{liu2016richer}, 
take advantage of its remarkable ability of 
 hierarchical feature learning and have demonstrated state-of-the-art F-score performance on the benchmarks such as BSDS500 \cite{arbelaez2011contour} and NYUDv2 \cite{Silberman:ECCV12}.

\begin{figure}[t]
\minipage{0.25\columnwidth}
\begin{subfigure}{\columnwidth}
\hrule
\includegraphics[width=\linewidth]{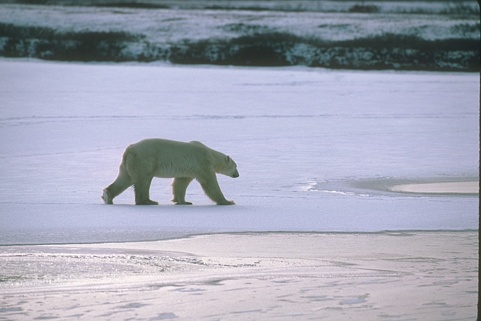}
\caption{}
\end{subfigure} 
\endminipage
\minipage{0.25\columnwidth}
\begin{subfigure}{\columnwidth}
\includegraphics[width=\linewidth]{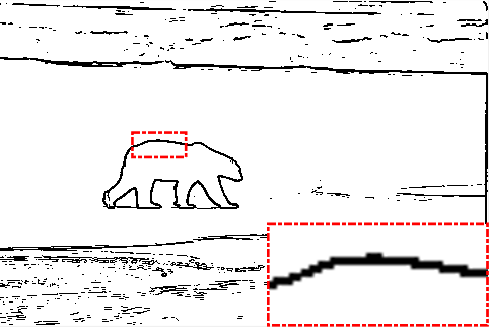}
\caption{}
\label{gt}
\end{subfigure} 
\endminipage
\minipage{0.25\columnwidth}
\begin{subfigure}{\columnwidth}
\includegraphics[width=\linewidth]{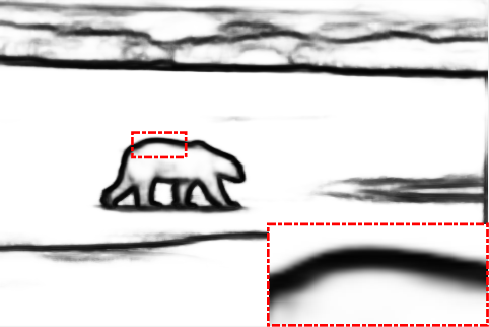}
\caption{}
\label{thick output}
\end{subfigure} 
\endminipage
\minipage{0.25\columnwidth}
\begin{subfigure}{\columnwidth}
\includegraphics[width=\linewidth]{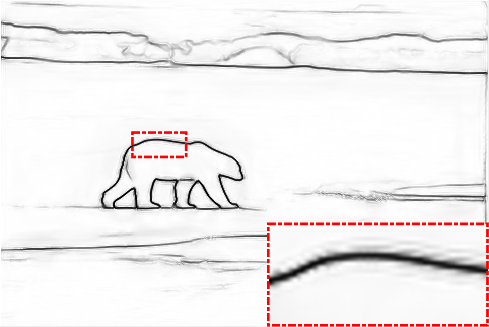}
\caption{}
\label{my_result}
\end{subfigure} 
\endminipage
\caption{ (a) is an example image from the BSDS500 dataset \cite{arbelaez2011contour}. 
(b) is the result of the Sobel detector \cite{sobel1970camera}. 
Here (c) is the output of the HED detector \cite{xie2015holistically}. (d) is the output of our proposed method. 
All the predictions do not apply post-processing.}
\label{img1}
\end{figure}

Although CNN-based methods are good at producing semantically meaningful contours, 
we observe a common behavior that their prediction is much thicker than the result of classic methods. 
For example, in Figure \ref{img1} we show two prediction examples  from  
the Sobel detector \cite{sobel1970camera} and the HED detector, respectively.  
The edge of the polar bear that we highlight in the dotted rectangle on the HED result is roughly 10 pixels wide, 
which is two times wider than the Sobel result (roughly 4 pixels). 
Note that, the behavior of thick prediction is not only on the result of HED but also can be found in many recent representative works 
such as RCF, Casenet \cite{yu2017casenet} and CEDN \cite{yang2016object}.

Existing works  in the literature seldom  discuss this issue of predicted boundaries being overly thick.
One possible reason %
is that 
edge detection methods typically apply edge thinning post-processing to obtain  one-pixel wide results after generating an initial prediction. 
Therefore it seems no difference that how wide the initial prediction is.  
However, this behavior attracts our attention and we believe it is worth  finding out the reason behind it 
which in turn improves the quality of prediction. 
The work in    \cite{wang2017deep}    addressed this problem by proposing a 
refinement architecture (encoder-decoder) for achieving crips edges. As we show in   our  experiments, 
 it only slightly improves the result of HED.
 Instead of modifying the convolutional network for boundary detection, we   address this problem by investigating the loss function.

In this work, we explore and solve the thickness issue of CNN-based boundary predictions. We present an end-to-end fully convolutional network 
which is accurate, fast and convenient to perform image-to-boundary  prediction. %
Our method consists of two key components,
 which are a fully convolutional neural network of the bottom-up/top-town architecture and a simple yet effective loss function. 
The method can automatically learn rich hierarchical features, resolve ambiguity in prediction and predict crisp results without postprocessing. 
Figure~\ref{img1} gives an example of the improvement of edge quality between our method and the HED detector. 
More examples can be found in Section~\ref{experiment}. 
We demonstrate that tackling the issue of thickness is critical for CNNs performing crisp edge detection, which improves the visual result 
as well as promotes the performance in terms of  boundary detection evaluation metrics. 
We achieve the state-of-the-art performance on the BSDS500 dataset with the ODS F-score of 0.815 and 
the fast version of our method achieves ODS F-score of 0.808 at the speed of  30 FPS.  

\section{Related work}

    Edge detection has been studied for over forty years. There are plenty of related works and here we only highlight a few representative works. 
Early edge detectors focus on computing the image gradients to obtain the edges \cite{kittler1983accuracy,canny1986computational,Fram1975On,perona1990scale}. 
For example, the Sobel detector \cite{kittler1983accuracy} slides a 3$\times$3 filter on a gray image to compute the image gradient for the response to the edge pixel. 
The  Canny detector \cite{canny1986computational} goes a step further by removing the noise on the output map and employing non-maximum suppression to extract one-pixel wide contour. 
These traditional methods are often used  as one of the fundamental features in many computer vision applications \cite{senthilkumaran2009edge,Lowe2004Distinctive,siddiqui2010human}. 
Learning-based methods \cite{martin2004learning,arbelaez2011contour,dollar2006supervised,dollar2015fast}  often  ensemble different low-level features and train a classifier to generate object-level contours. 
Although these methods achieved great performance compared to  traditional methods, they rely on hand-crafted  features which limit their room for improvement.

Recent state-of-the-art methods for edge detection \cite{xie2015holistically,liu2016richer,kokkinos2015pushing} are built on deep convolutional neural networks \cite{krizhevsky2012imagenet,lecun1990handwritten}. 
CNNs have the virtue of automatically learning low-level, middle-level and high-level features from training samples. 
Taking the advantage, CNN-based edge detectors can seek high response of edge on the fused feature map instead of the original image, which helps them generate semantic meaningful contour and achieve remarkable performance. 
To be more specific, DeepEdge \cite{bertasius2015deepedge} extracts multiple patches surrounding an edge candidate point (extracted by the Canny detector) and feeds these patches into a multi-scale CNN to decide if it is an edge pixel. 
DeepContour \cite{shen2015deepcontour} is also a patch-based approach which first divides an image into many patches then put these patches into the network to detect if the patch has a contour. 
Differing from these works, the HED detector \cite{xie2015holistically} is an end-to-end fully convolutional neural network which takes an image as input and directly outputs the prediction. 
It proposes a weighted cross entropy loss and takes the skip-layer structure to make independent predictions from each block of the pre-trained VGG model \cite{simonyan2014very} and average the results. 
RCF \cite{liu2016richer} also utilizes the skip-layer structure and the similar loss with HED, yet it makes independent predictions from each convolutional layer of the VGG model. 
CEDN \cite{yang2016object} employs an encoder-decoder network and train the network on the extra data of Pascal VOC dataset. 
CASENet \cite{yu2017casenet} propose a novel task which is to assign each edge pixel to one or more semantic classes and solve the task by utilizing an end-to-end system similar to HED and RCF.

Summarizing  the development of deep learning based methods, we find that the HED detector is very popular and has enlightened many subsequent methods such as RCF, CASENet and the works mentioned in the paper \cite{xie2017holistically}. 
However, we observe empirically  that the weighted cross entropy loss employed  by the HED detector may have contributed to the resulted edges  being thick. 
We verify this  in the next section.

\textbf{Contributions} In this work, we develop an end-to-end edge detection method.  
    Our main contributions are as follows. 
    We aim to detect crisp boundaries in images using deep learning.
    We explore the  issue of predicted edges being overly thick, which can be found in almost all  recent CNNs based methods.
    We propose a method that  manages to tackle the thickness issue. It allows CNN-based methods to predict crisp edge without postprocessing.
    Furthermore, our experiments show that our method outperforms previous state-of-the-art methods on the BSDS500 and NYUDv2 datasets.

\section{The proposed method}

In this section, we describe the details of the proposed method. 
Loss function is the most important component in an end-to-end dense prediction system 
since the quality of prediction is most affected by its loss. 
Thus we first revisit the weighted cross entropy loss used in previous state-of-the-art methods. 
We then propose our edge detection system, including the loss based on image similarity and the network of bottom-up/top-down structure.

\begin{figure}[t]
\centering
\minipage{0.33\columnwidth}
\begin{subfigure}{\columnwidth}
\includegraphics[width=\linewidth]{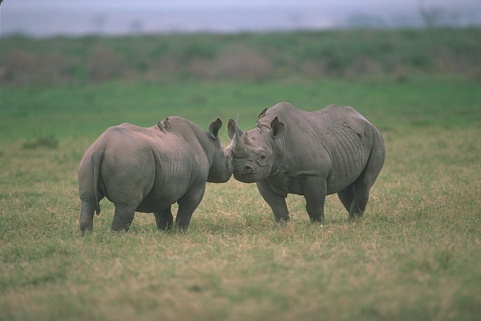}
\caption{}
\end{subfigure} 
\endminipage
\minipage{0.33\columnwidth}
\begin{subfigure}{\columnwidth}
\includegraphics[width=\linewidth]{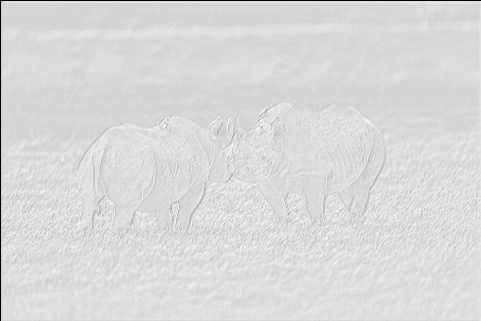}
\caption{ }
\label{without_weight}
\end{subfigure} 
\endminipage
\minipage{0.33\columnwidth}
\begin{subfigure}{\columnwidth}
\includegraphics[width=\linewidth]{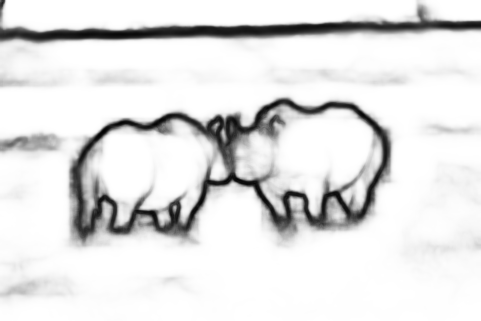}
\caption{}
\label{with_weight}
\end{subfigure} 
\endminipage
\vspace{-.2cm}
\caption{
     A  simple test on the class-balance weight $\beta$. From left to right: (a) is the original image from the BSDS500 dataset. 
     (b) is the result of using the standard  cross-entropy loss, i.e.,  $\beta = 0.5$. 
     (c) is the result of using the weighted cross-entropy loss. }
\label{contrast test}
\end{figure}

\subsection{Revisiting  the weighted cross-entropy loss for edge detection}
Previously  fully convolutional networks (FCN) based 
edge-detection methods often employ  the weighted cross entropy loss as adopted  by the HED detector. 
It is well known that the cross-entropy loss is used to solve two-class classification.
 However, the edge/non-edge pixels are of a highly imbalanced  distribution (the majority of pixels are non-edge) thus the direct use of  the cross-entropy loss would fail to train the network. 
To tackle the issue, HED uses  the weighted cross entropy loss,  which writes 
\begin{equation}
\label{hed loss}
\begin{aligned}
\mathcal{L}(W,w) = -\beta\sum_{j \in Y_{+}}\log\Pr(y_{j}=1 | X;W,w) 
                                   -(1-\beta )\sum_{j \in Y_{-}}\log\Pr(y_{j}=0 | X;W,w),                                 
\end{aligned}
\end{equation}
where $Y_{+}$ and $Y_{-}$ denotes the 
edge pixel set and non-edge pixel sets, respectively. $\beta = |Y_{-}|/|Y|$ and $1-\beta = |Y_{+}|/|Y|$. $X$ is the input image and $\Pr(y_{j}| X;W,w)$ is computed 
using softmax  on the classification scores at pixel $y_{j}$.

The class-balance weights $\beta$ and $1-\beta$ are used to preserve and suppress the losses from the class of edge pixels and non-edge pixels, respectively. 
This simple solution helps CNN manage to better train the network. 
We perform a comparison  test   that     uses the standard  cross-entropy loss and the weighted loss on the same HED network structure,  demonstrating 
the effectiveness of the weight $\beta$. The result of the test is shown in Figure~\ref{contrast test}.
As we can see, the standard  loss fails to train the network since the result (Figure~\ref{without_weight})  
    is not an edge map but an `embossed' image. 
    However, if we carefully look at its details,     we are able to find the contour of rhinoceros that has  reasonable thickness and is thinner 
    than the results  of the weighted loss (Figure~\ref{with_weight}). 
It is likely to indicate that,  although the class-balance weights $\beta$ and $1-\beta$ manage to make CNNs successfully trained, they cause the `thickness' issue. %
This finding explains why recent methods such as RCF and Casenet tend to output overly thick edges.   These two methods have employed  the cross-entropy loss with the same strategy, 
i.e.,  setting weights on edge/non-edge pixels to balance the loss. 
To make the network trainable and output  a crisp  prediction at the same time, we will need alternative solutions.

\subsection{The proposed loss function for edge detection}
\label{loss}

We have shown   that a distinct characteristic of edge map is that the data is highly biased because the vast majority of the pixels are non-edges. 
This highly biased issue would cause the learning to %
fail to find the crisp edges which are the `rare events'. 

Similar to our task,  many  applications such as fraud detection, medical image processing, and text classification are dealing with class imbalance data and 
there are corresponding solutions to these tasks \cite{gu2007making,tang2005bias,lusa2010class,haider2014unconscious,phillips2009generative}. 
Inspired by the work of \cite{milletari2016v} using the Dice coefficient \cite{dice1945measures} to solve the class-imbalance problem, we %
propose to use the Dice coefficient for edge detection.

    Given an input image $I$ and the ground-truth $G$, the activation map $M$ is the input image $I$ processed by a fully convolutional network $F$. 
Our objective is to obtain a prediction $P$. 
Our loss function $L$ is given by
\begin{equation}
\label{our loss}
L(P,G) = Dist(P,G) = \frac{\sum_{i}^{N}p_{i}^{2} + \sum_{i}^{N}g_{i}^{2} }{2\sum_{i}^{N}p_{i}g_{i}},
\end{equation}
where $p_{i},g_{i}$ denote the value of $i$-th pixel on the prediction map $P$ and the ground-truth $G$, respectively. 
The prediction map $P$ is computed from the activation map $M$ by the sigmoid function. 
The loss function $L$ is the reciprocal of the Dice coefficient. Since the Dice coefficient is a measure of similarity of two sets. 
Our loss is to compare the similarity of two sets $P$, $G$ and minimizes their distance on  the training data. 
We do not need to consider the issue of balancing the loss of edge/non-edge pixels by using the proposed loss and are able to achieve our 
 objective---make the network trainable and predict  crisp   edges at the same time.

We should emphasize the way of computing our total loss in a mini-batch. Given a mini-batch of training samples and their corresponding ground-truth, our total loss is given by 
\begin{equation}
\label{our loss in a minibatch}
L(MP, MG) = \sum_{i}^{M} Dist(MP_{i}, MG_{i}), 
\end{equation}
where $MP$ and $MG$ denote a mini-batch of predictions and their ground-truth, respectively. 
$M$ is the total number of training samples in the mini-batch. 
Since our loss function is based on the similarity of per image-ground-truth pair, our total loss of a mini-batch is the sum of the total distances  over   all pairs. 

To achieve better performance, we propose to combine the cross-entropy loss and the proposed Dice loss. The Dice  loss may be thought of as 
 image-level  that  focuses on the similarity of two sets of image pixels. 
The cross-entropy loss concentrates on the pixel-level difference, since it is the sum of the distance of every corresponding pixel-pair between prediction and the ground-truth. 
Therefore the combined loss is able to hierarchically minimize the distance from image-level to pixel-level. 

Our final loss function is given by: %
\begin{equation}
\label{basic loss}
L_{\text{final}}(P,G) =  \alpha L(P,G) + \beta L_{c}(P,G),
\end{equation}
where $L(P,G)$ is  Equation \ref{our loss}; 
$L_{c}(P,G)$ is the normal cross-entropy loss which is $L_{c} (P,G)= -\sum_{j}^{N}(g_{j}\log p_{j} + (1-g_{j})(1-\log p_{j}) )$. 
$N$ is the total pixel number of an image. 
$\alpha$ and $\beta$ are the parameters to control the influence of two losses. 
In experiments we set $\alpha = 1$ and $\beta = 0.001$. 
We also tried to use the weighted cross-entropy loss (Equation \ref{hed loss}) instead of $L_{c}$, and no improvement is  observed. 
To compute the total loss in a mini-batch, we use Equation \ref{our loss in a minibatch} where $Dist(P,G)$ is replaced by $L_{\text{final}}(P,G)$. 
We emphasize that the proposed Dice  loss $L(P,G)$ is the cornerstone for generating crisp edges. 
Using only the  proposed Dice  loss, we achieve an  ODS F-score of .805 on the BSDS500 dataset.

The formulation \eqref{basic loss} can be differentiated yielding the gradient
\begin{equation}
\label{gradient}
\frac{\partial L_{\text{final}}}{\partial p_k} = \alpha \frac{2p_{k}\sum_{i=1}^{N}p_{i}g_{i} - g_{k}(\sum_{i=1}^{N}p_{i}^2 + \sum_{i=1}^{N}g_{i}^2)}{2(\sum_{i=1}^{N}p_{i}g_{i})^2} - \beta \frac{2g_{k}-1}{p_{k}}
\end{equation}
computed with respect to the $k$-th pixel of the prediction.

In the next subsection, we  describe our network structure.
\begin{figure}[t!]
\centering
\includegraphics[height=90mm]{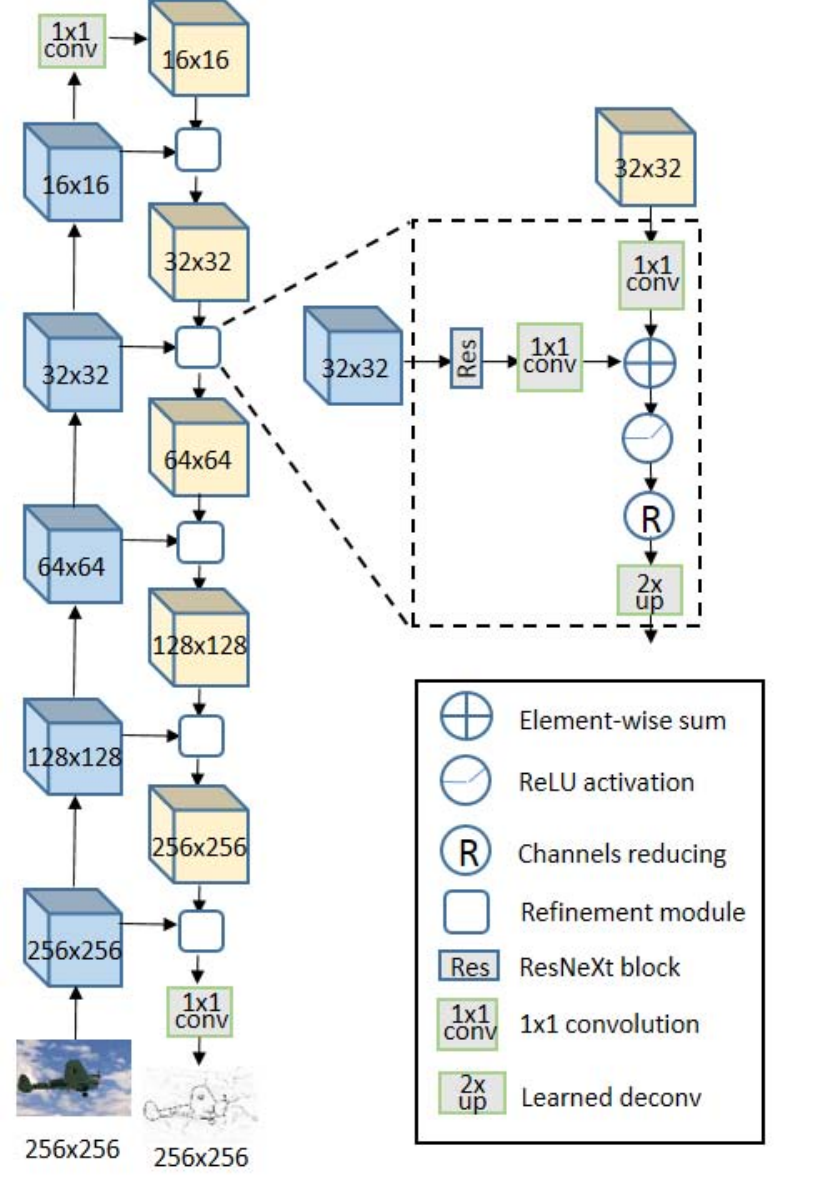}
\caption{Overview of the proposed network. Blue cubics indicate the features on the bottom-up path while yellow cubics indicate the mask-encoding on the top-down path. 
The backbone of our network is the VGG-16 model in which the last pooling layer and all the fully connected layers are removed. 
The mask-encoding from conv5\_3 layer  repeatedly goes  through the proposed refinement module to recover its resolution. 
In a refinement module, the mask-encoding  is 
 fused with the side-output features and then reduces its channels by a factor of 2 and double its resolution to prepare for the fusion in the next refinement module.}%
\label{structure fig}
\end{figure}
\subsection{Network architecture} 
\label{structure}
We attempt   to select the network structure which 
has multiple stages to efficiently capture hierarchical features  
and is able to fuse the features of different levels, so as to generate semantically  meaningful contours. 
The success of HED shows the great importance of a carefully designed structure. 
In this paper, we look at another advanced structure, that is the bottom-up/top-down architecture \cite{pinheiro2016learning} 
for inspiration to make better use of hierarchical features.
 The method of \cite{pinheiro2016learning} achieves improved accuracy of object segmentation by proposing a novel top-down refinement approach. 
 We hypothesize that this structure may also work for edge detection well 
 since our task is related to object segmentation. To our knowledge, we may be  the first to adopt the bottom-up/top-down structure for edge detection.

We follow the setting of the network \cite{pinheiro2016learning} 
to apply the VGG-16 model \cite{simonyan2014very} as the backbone and stack its 
`refactored' structure of the refinement module to recover the resolution of features. 
However, we have the following modifications at the refinement module to make it suitable for edge detection: 
(i) to better extract side feature from each stage of VGG-16, we use  the \emph{ResNeXt} \cite{Xie2016Aggregated} block
 to connect each side output layer, respectively conv1\_2, conv2\_2, conv3\_3, conv4\_3 and conv5\_3. Thus, the feature from each side output 
  first goes  through a \emph{ResNeXt} block then is  fused with the mask-encoding from the top-down path; 
(ii) we use $1\times1$ \emph{conv} layer to replace original $3\times3$ \emph{conv} layers of the module. 
By doing so, we find the performance is improved with the decrease of model complexity; 
(iii) we use the learned \emph{deconv} layer to double the resolution of fused features.
 Especially, the \emph{deconv} layer is grouped. The group number equals to the channel number of the fused features. 
The grouped \emph{deconv} layer allows our model to keep the performance with less model complexity. 
The modified refinement module is fully back-propable. We show the overall structure in Figure~\ref{structure fig} and our refinement module in the dotted rectangle.

Our network is simple yet very effective for edge detection. 
We  highlight  that it is vital for an edge-detection network to increase the ability of feature extraction with the decrease of model complexity. 
Compared to the original structure, our network has the advantage of using fewer parameters to achieve 
better performance. To be more specific, our network has 15.69M parameters and achieves an ODS of .808 on the BSDS500 dataset. 
Without the modifications described in  (ii) and (iii), the parameter number increases to 22.64M but the performance decreases to ODS of .802. 

 The  reason behind this phenomenon might be due to  overfitting,  as 
the dataset for edge detection has limited number of training samples (for example, the BSDS500 dataset only has only 200 training images). 
In experiments, we tried a few  more sophisticated   bottom-up/top-down networks such as Refinenet \cite{lin2017refinenet}, 
but failed to achieve better performance possibly because of limited training data. Using the \emph{ResNeXt} block is for the same reason. 
It groups the inside \emph{conv} layers to decrease the model complexity. 
We also test the \emph{ResNet} block \cite{he2016deep} to extract the side features, 
which is used to compare the performance against  the \emph{ResNeXt} block. 
We find that they are both helpful to boost the performance while the \emph{ResNeXT} block performs slightly better with roughly  50\% complexity of the \emph{ResNet} block.

\section{Experiments}
\label{experiment}
In this section, we first present the implementation details as well as a brief description of the   datasets. 
Our experiments start with an ablation study of the proposed method. 
We then conduct a comparative study on HED to demonstrate the effectiveness of the proposed loss on the crispness of prediction. 
We further compare our method with the state-of-the-art edge detectors and demonstrate the advantages.

\subsection{Implementation details}
We implement our method using  Pytorch \cite{AdamPaszke}. 
We evaluate edge detectors on 
Berkeley Segmentation Dataset (BSDS 500) and 
NYU depth dataset (NYUD), 
which are widely used in previous works \cite{bertasius2015deepedge,shen2015deepcontour,xie2015holistically,liu2016richer,yang2016object}. 
The hyper-parameters of our model include: %
mini-batch size (36), input image resolution ($480\times320$), weight-decay (1e$-4$), 
 training epochs  (30). 
We use the ADAM solver \cite{DBLP:journals/corr/KingmaB14} for optimization. 

\begin{figure*}[t]
\centering
\begin{subfigure}{0.135\textwidth}
{\includegraphics[scale = 0.15]{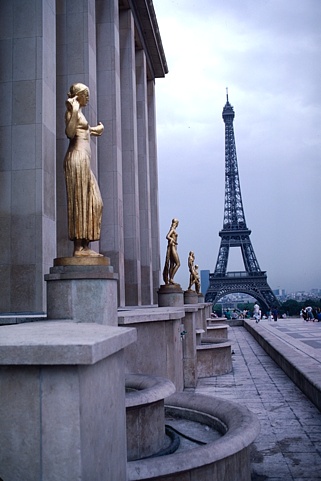}}
\end{subfigure}
\begin{subfigure}{0.135\textwidth}
{\includegraphics[scale = 0.15]{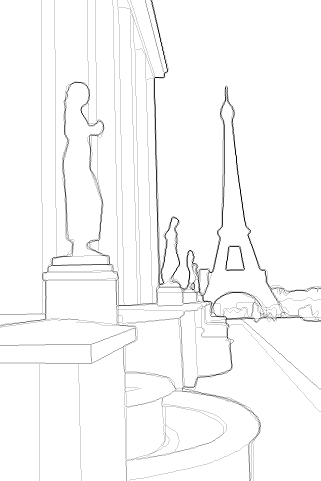}}
\end{subfigure}
\begin{subfigure}{0.135\textwidth}
{\includegraphics[scale = 0.15]{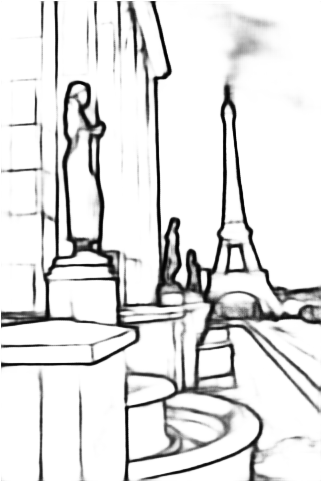}}
\end{subfigure}
\begin{subfigure}{0.135\textwidth}
{\includegraphics[scale = 0.15]{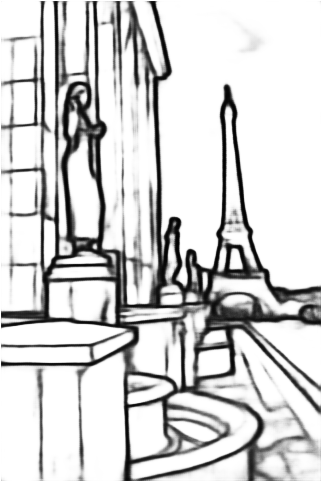}}
\end{subfigure}
\begin{subfigure}{0.135\textwidth}
{\includegraphics[scale = 0.15]{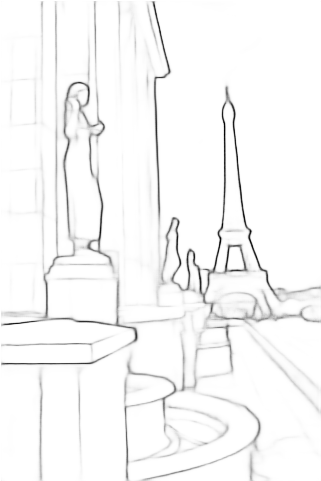}}
\end{subfigure}
\begin{subfigure}{0.135\textwidth}
{\includegraphics[scale = 0.15]{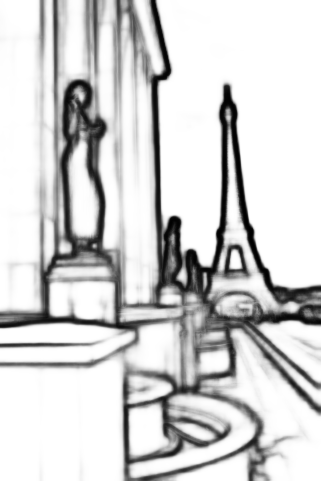}}
\end{subfigure}
\begin{subfigure}{0.135\textwidth}
{\includegraphics[scale = 0.15]{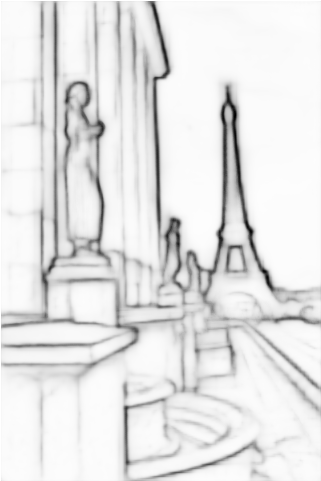}}
\end{subfigure}
\vskip 0.05in
\begin{subfigure}{0.135\textwidth}
{\includegraphics[scale = 0.15]{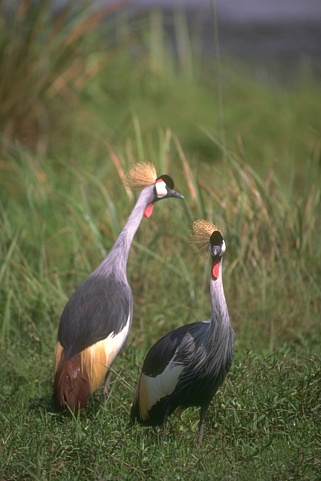}}
\caption{} 
\end{subfigure}
\begin{subfigure}{0.135\textwidth}
{\includegraphics[scale = 0.15]{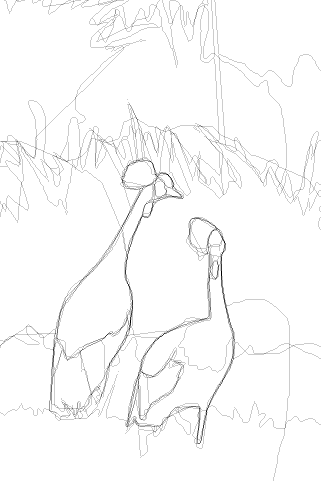}}
\caption{} 
\end{subfigure}
\begin{subfigure}{0.135\textwidth}
{\includegraphics[scale = 0.15]{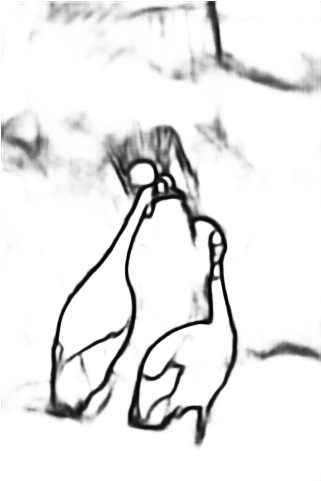}}
\caption{} 
\end{subfigure}
\begin{subfigure}{0.135\textwidth}
{\includegraphics[scale = 0.15]{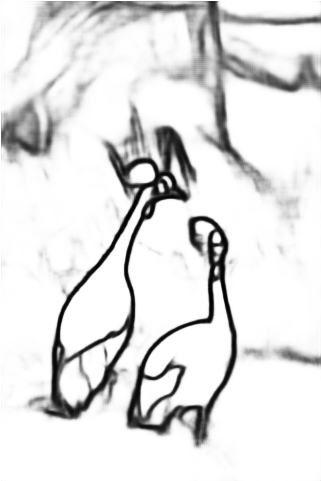}}
\caption{} 
\end{subfigure}
\begin{subfigure}{0.135\textwidth}
{\includegraphics[scale = 0.15]{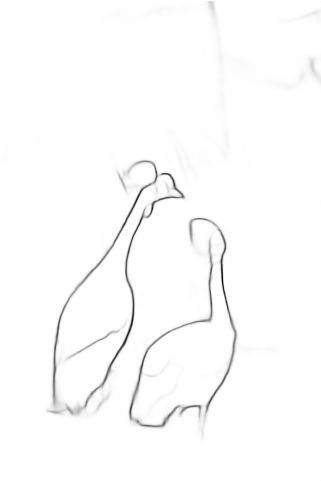}}
\caption{} 
\end{subfigure}
\begin{subfigure}{0.135\textwidth}
{\includegraphics[scale = 0.15]{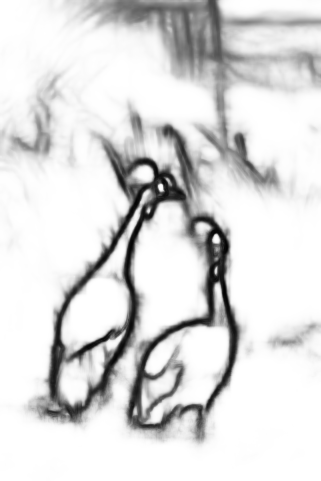}}
\caption{} 
\end{subfigure}
\begin{subfigure}{0.135\textwidth}
{\includegraphics[scale = 0.15]{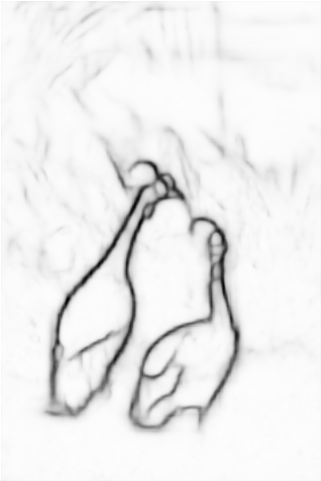}}
\caption{} 
\end{subfigure}
\caption{Illustration of the qualitative results of the ablation study as well as  applying the proposed loss with  HED. 
From left to right: (a) input images in the BSDS500 dataset; 
(b) ground-truth; 
(c),(d),(e) are the predictions of the methods \emph{Ours-w/o-rN-w/o-FL, 
Ours-w/o-FL} and \emph{Ours} in the ablation study, respectively; 
(f), (g) are the predictions of the methods \emph{HED-BL} and \emph{HED-FL} in the comparative study, respectively. 
Our method, especially the proposed loss, shows a clear advantage in generating sharp boundaries.
}
\label{figs of ablation study}
\end{figure*}

Beside the hyper-parameters, the following  several key issues  are worth mentioning:\\ 
\textbf{Data augmentation} 
\label{data aug}
Data augmentation is an effective way to boost performance when the amount of training data is limited. 
We first random scale the image-label pairs (0.7 to 1.3). 
We then rotate the pairs to 16 different angles and crop the largest rectangle in the rotated angle. 
We finally flip the cropped images, 
which leads to an augmented training set from 200 images to more than 100k images.\\
\textbf{Up-sampling method} 
We employ learned deconvolution in the backward-refining pathway to 
progressively increase the resolution of feature maps. 
Although bilinear upsampling was demonstrated to be  useful in HED, 
it is abandoned in our method. 
 We observe in experiments that bilinear upsampling may make  the prediction discontinuous at a number of locations 
 and cause a  slight decrease in performance. \\
\textbf{Multi-scale edge detection} 
Inspired by the works \cite{liu2016richer,wang2017deep},  during testing 
we make use of multi-scale edge detection to further improve performance. 
We first resize an input image to three different resolutions ($0.5\times$, $1.0\times$ and $1.5\times$ of the original size), 
which are fed into the network. We then resize the outputs back to the original size and average them to obtain   the final prediction. 

\subsection{BSDS500 dataset}
Berkeley Segmentation Dataset (BSDS 500) \cite{arbelaez2011contour} contains 200 training images, 100 validation images, and 200 testing images. 
Each image is annotated by multiple users. 
We use the train set (200 training images) for training 
and employ all the ground-truth labels to prepare the training data. 
That is, if an image has five annotations, we first create five copies of the image. 
Each copy is corresponding to one of the annotations, respectively. 
We then apply these five image-annotation pairs for data augmentation. 
This would introduce ambiguity in the ground-truth pairs because
 different annotators may disagree with each other  for   a small number of pixels. 
However, in this case, we are able to get more annotations for data augmentation. 
In the meantime, we observed that introducing certain ambiguity prevented the training from overfitting. 

\begin{table}[t]
\renewcommand{\arraystretch}{1.05}
\caption{Ablation studies  of the proposed method on the BSDS500 dataset. NMS stands for non-maximum suppression. 
`Ours-w/o-FL' refers to our method without the fusion loss. 
`w/o-rN' refers to without all the ResNeXt blocks in the backward-refining path.
}
\centering
\begin{tabular}{ c|c|c  }
 \hline
 Method                                                       &                 ODS (after/before NMS)         &              OIS (after/before NMS) \\
 \hline
Ours-w/o-rN-w/o-FL     &                    .797 / .671                                                    &               .815 / .678     \\
Ours-w/o-FL                    &                   .798 / .674                                                     &              .815 / .679      \\
Ours                                  &                   .800 /.693                                                     &               .816 /.700   \\ 
\hline
\end{tabular}
\label{ablation study table}
\end{table}

\begin{table}[t]
\renewcommand{\arraystretch}{1.05}
\caption{Comparative studies  of HED. HED-BL refers to HED trained by the balanced cross-entropy loss. 
HED-FL refers to HED trained with  the proposed fusion loss.}
\centering
\begin{tabular}{ c|c|c  }
 \hline
 Method                    &                 ODS (after/before NMS)         &              OIS (after/before NMS) \\
 \hline
HED-BL           &                    .781 / .583                                                    &               .798/ .598     \\
HED-FL           &                    .783 / .635                                                     &              .802 / .644      \\
\hline
\end{tabular}
\label{comparative study}
\end{table}

\subsubsection{Ablation study}
We first conduct a series of ablation studies to 
evaluate the importance of each component in the proposed method.  
Our first experiment is to examine the effectiveness of 
the basic encoder-decoder network (Ours-w/o-rN-w/o-FL) for the task. 
To this end, our baseline model is 
the proposed network removing all the ResNeXt blocks in the backward-refining path.
We train this baseline using the balanced cross-entropy loss.
Moreover, we trained two versions of the proposed network 
via the balanced cross-entropy loss (Ours-w/o-FL) and the proposed fusion loss (Ours), respectively.

The accuracy of prediction is evaluated via two standard measures: 
fixed contour threshold (ODS) and per-image best threshold (OIS). 

Previous works tend to only examine the correctness of prediction 
since they apply  a standard non-maximal suppression (NMS) to predicted edge maps before evaluation.
While in this study and the following comparative study, 
we would like to evaluate each model twice (before and after NMS). 
By doing so, we can examine both the correctness and the sharpness since low-crispness 
prediction is prone to achieve low ODS scores, 
without the aid of NMS. 
We are aware that CED \cite{wang2017deep} and PMI \cite{isola2014crisp} 
apply  a different way to benchmark the crispness of predictions 
by varying a matching distance parameter. 
However, we consider that  directly evaluating non-NMS results 
is simpler yet effective for the same purpose.

The quantitative results are listed in Table \ref{ablation study table} 
and two qualitative examples are shown in Figure \ref{figs of ablation study}(c), (d) and (e). 
From the results, we  observe  three findings. 
Firstly, each component is able to improve performance; 
Secondly, a convolutional encoder-decoder network may be more competent for  the task, 
compared to the network of HED. 
We can see that the baseline (Ours-w/o-rN-w/o-FL) achieves an  ODS score .797, 
which significantly outperforms HED (.790 on the BSDS500 dataset).
Lastly, both the quantitative and the qualitative results have demonstrated 
the effectiveness of the proposed fusion loss. 
By  simply  using  the proposed fusion  loss, the ODS f-score (before NMS) of our network is increased from .674 to .693
and the improvement of boundary sharpness can be also 
observed in Figure \ref{figs of ablation study}(d) and (e).

\subsubsection{Improving the crispness of HED}
As mentioned in  Section \ref{loss}, 
the proposed fusion loss plays a key role in our method in terms  of generating   sharp boundaries, 
which was demonstrated in the ablation study. 
One may ask a question: 
\emph{Does the fusion loss only work on the convolutional encoder-decoder network? 
Could it also allow different methods such as HED to improve crispness? }

To answer this question, we perform a comparative study on the HED edge detector. 
Similar to the ablation experiments, we evaluate two versions of HED: 
one is trained by means of the proposed fusion loss, 
the other is applying the balanced cross-entropy loss. 
Both  methods are trained using deep supervision. 
Note that our training data of BSDS500 is generated in a different way, compared to HED \cite{xie2017holistically}, 
thus the performance of the re-implemented HED is slightly different from the original paper. 
We summarize the quantitative results in Table \ref{comparative study} and 
show two qualitative examples in Figure \ref{figs of ablation study}(f) and (g).

The results are consistent with those of the ablation experiments. With the use  of the proposed loss, 
\emph{HED-FL} improves the non-NMS results over \emph{HED-BL} by almost 9\%, 
which is a significant increase at boundary crispness.

\begin{figure*}[t]
\centering
\begin{minipage}{0.19\textwidth}
{\includegraphics[width=\linewidth,height=1.2cm]{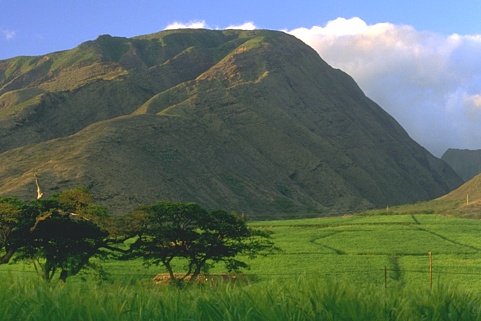}}
\end{minipage}
\begin{minipage}{0.19\textwidth}
{\includegraphics[width=\linewidth,height=1.2cm]{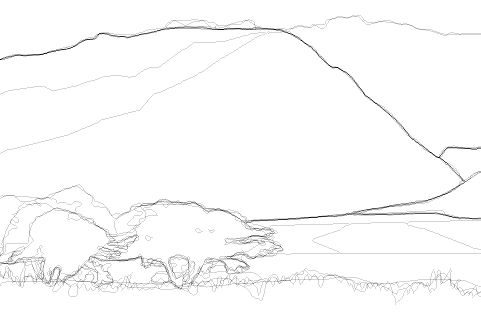}}
\end{minipage}
\begin{minipage}{0.19\textwidth}
{\includegraphics[width=\linewidth,height=1.2cm]{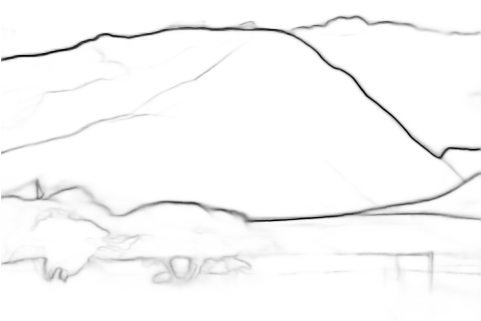}}
\end{minipage}
\begin{minipage}{0.19\textwidth}
{\includegraphics[width=\linewidth,height=1.2cm]{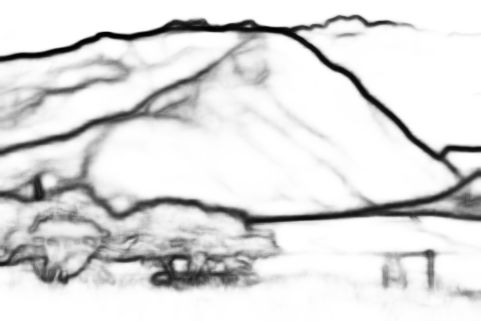}}
\end{minipage}
\begin{minipage}{0.19\textwidth}
{\includegraphics[width=\linewidth,height=1.2cm]{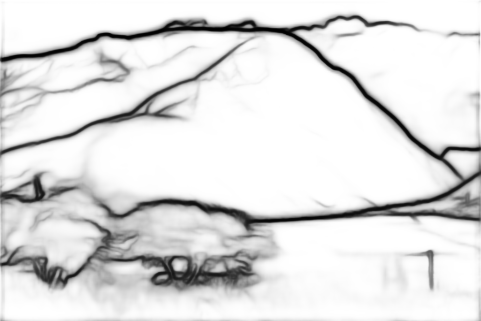}}
\end{minipage}
\vskip 0.05in
\begin{minipage}{0.19\textwidth}
{\includegraphics[width=\linewidth,height=1.2cm]{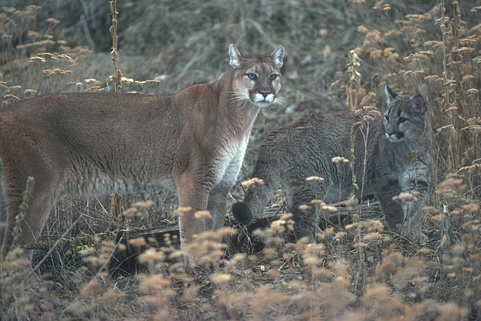}}
\end{minipage}
\begin{minipage}{0.19\textwidth}
{\includegraphics[width=\linewidth,height=1.2cm]{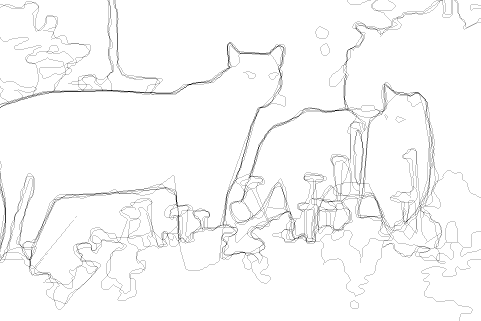}}
\end{minipage}
\begin{minipage}{0.19\textwidth}
{\includegraphics[width=\linewidth,height=1.2cm]{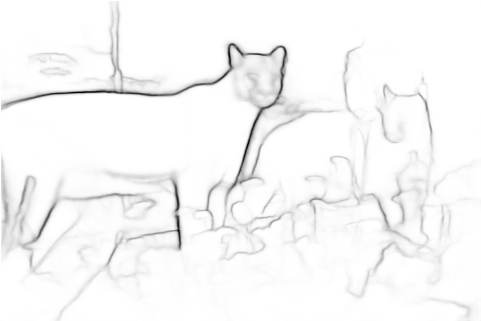}}
\end{minipage}
\begin{minipage}{0.19\textwidth}
{\includegraphics[width=\linewidth,height=1.2cm]{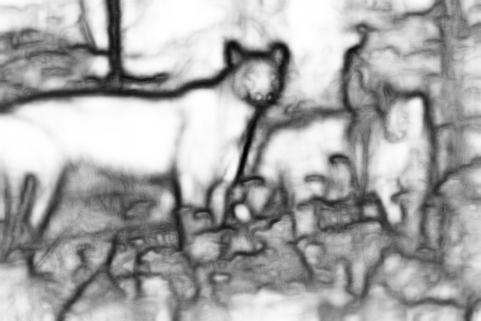}}
\end{minipage}
\begin{minipage}{0.19\textwidth}
{\includegraphics[width=\linewidth,height=1.2cm]{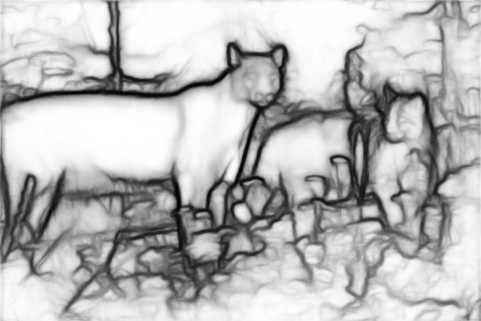}}
\end{minipage}
\vskip 0.05in
\begin{minipage}{0.19\textwidth}
{\includegraphics[width=\linewidth,height=1.2cm]{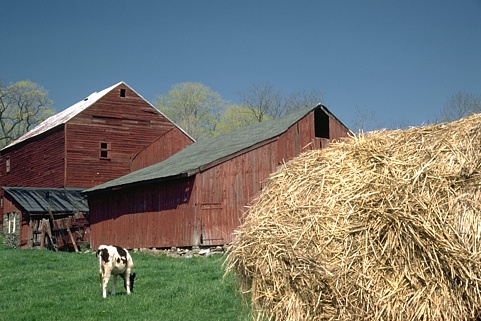}}
\end{minipage}
\begin{minipage}{0.19\textwidth}
{\includegraphics[width=\linewidth,height=1.2cm]{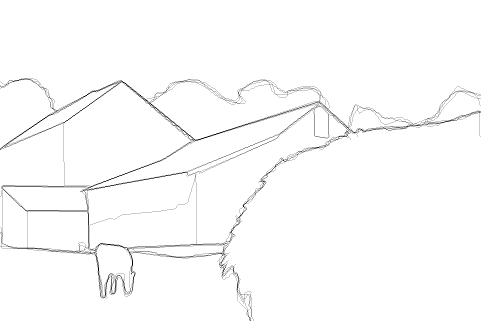}}
\end{minipage}
\begin{minipage}{0.19\textwidth}
{\includegraphics[width=\linewidth,height=1.2cm]{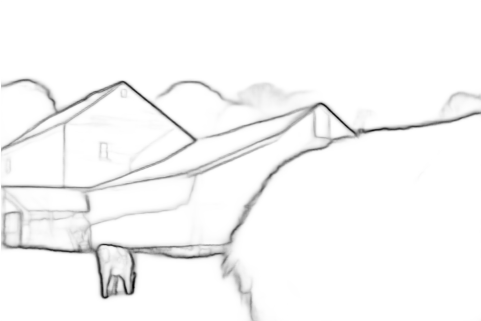}}
\end{minipage}
\begin{minipage}{0.19\textwidth}
{\includegraphics[width=\linewidth,height=1.2cm]{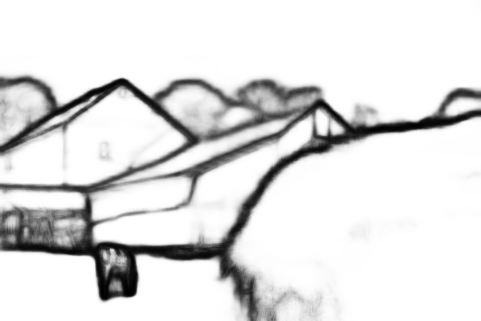}}
\end{minipage}
\begin{minipage}{0.19\textwidth}
{\includegraphics[width=\linewidth,height=1.2cm]{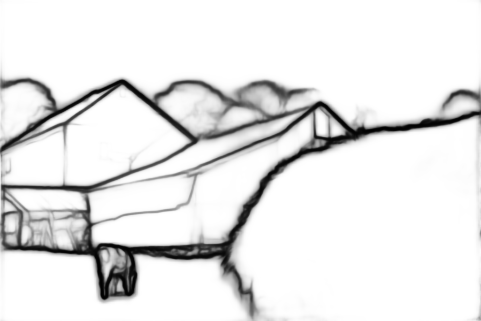}}
\end{minipage}
\vskip 0.05in
\begin{subfigure}{0.19\textwidth}
{\label{main:a}\includegraphics[width=\linewidth,height=1.2cm]{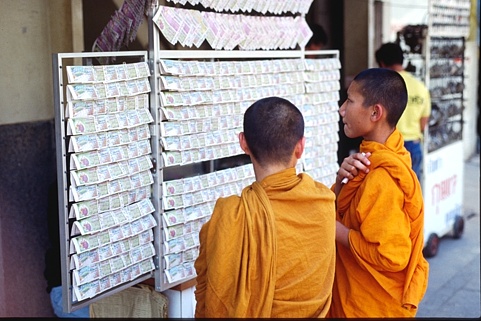}}
\caption{Input Image}
\end{subfigure}
\begin{subfigure}{0.19\textwidth}
{\label{main:e}\includegraphics[width=\linewidth,height=1.2cm]{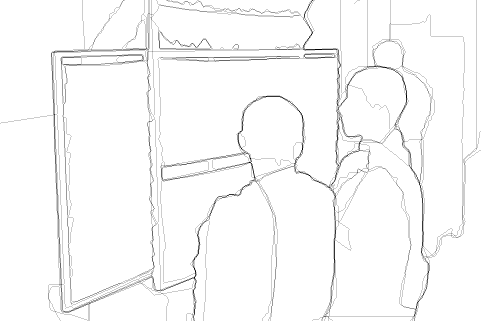}}
\caption{GT}
\end{subfigure}
\begin{subfigure}{0.19\textwidth}
{\label{main:b}\includegraphics[width=\linewidth,height=1.2cm]{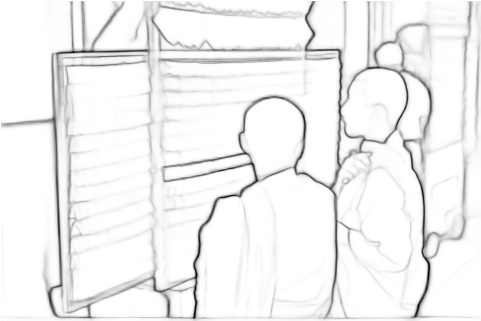}}
\caption{Ours}
\end{subfigure}
\begin{subfigure}{0.19\textwidth}
{\label{main:c}\includegraphics[width=\linewidth,height=1.2cm]{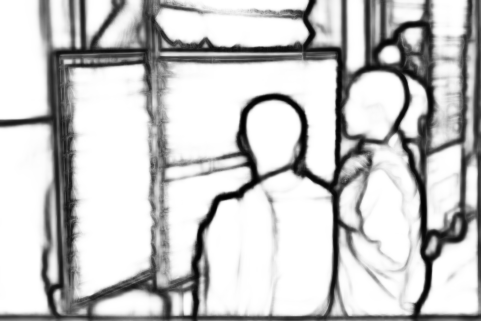}}
\caption{RCF \cite{liu2016richer} }
\end{subfigure}
\begin{subfigure}{0.19\textwidth}
{\label{main:d}\includegraphics[width=\linewidth,height=1.2cm]{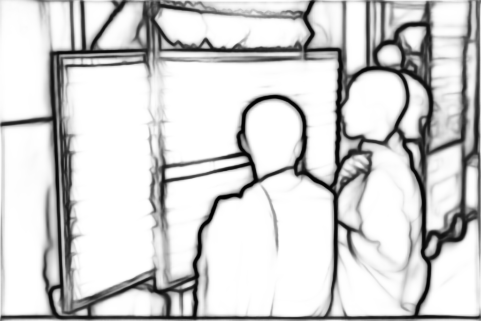}}
\caption{CED \cite{wang2017deep}}
\end{subfigure}
\caption{State-of-the-art comparisons on BSDS500. From left to right:
 (a) the original images, (b)  ground-truth, (c) the predictions of the proposed method, (d) the results of the RCF detector, (e) the results of the CED detector. Note that all the predictions are ent-to-end outputs and not postprocessed.}
\label{figs of comparison on BSDS}
\end{figure*}

\subsubsection{State-of-the-art comparisons} 
In this subsection, we further compare against the top performing edge detectors. 
The methods to be evaluated are composed of two classes: 
the first class is deep-learning based, which includes HED \cite{xie2015holistically}, RCF \cite{liu2016richer}, DeepContour \cite{shen2015deepcontour}, DeepEdge \cite{bertasius2015deepedge}, CED \cite{wang2017deep}, 
HFL \cite{bertasius2015high}, CEDN \cite{yang2016object}, MIL+G-DSN+MS+NCuts \cite{kokkinos2015pushing} and our method; 
the second class contains SE  \cite{dollar2015fast}, gPb-UCM \cite{arbelaez2011contour} and the Canny detector \cite{canny1986computational}. 
We also follow the works of \cite{yang2016object,liu2016richer,kokkinos2015pushing,wang2017deep} to employ the extra training data from PASCAL VOC Context dataset \cite{mottaghi2014role}. 
The results are shown in Figure \ref{figs of comparison on BSDS}, Figure \ref{bsds_curve} and Table \ref{bsds numerical comparison}. 
\begin{table}[t]
\renewcommand{\arraystretch}{1.05}
\caption{Results on the BSDS500 dataset. 
MS refers to the multi-scale testing. 
VOC-aug refers to training with extra PASCAL VOC context data. $\dagger$ refers to GPU time.}
\centering
\begin{tabular}{ c|c|c|c  }
 \hline
 Method                                                             &                  ODS          &              OIS      &                FPS \\
 \hline
Canny \cite{canny1986computational}  &                  .611            &             .676      &                28\\
 gPb-UCM \cite{arbelaez2011contour}  &                   .729           &              .755      &                1/240\\
 SE  \cite{dollar2015fast}                            &                    .743           &                .763     &                2.5   \\
 \hline
 DeepContour \cite{shen2015deepcontour} &        .757           &                 .776      &                $1/30^{\dagger}$\\
 DeepEdge \cite{bertasius2015deepedge} &        .753           &                .772       &                $1/1000^{\dagger}$\\
 HFL \cite{bertasius2015high}                            &        .767           &                .788       &                 $5/6^{\dagger}$\\
 HED \cite{xie2015holistically}                           &        .788           &                .808       &                  $30^{\dagger}$\\
 CEDN \cite{yang2016object}                            &        .788           &                .804       &                  $10$                       \\
 MIL+G-DSN+MS+NCuts \cite{kokkinos2015pushing} &   .813     &        .831        &                       1                        \\
 RCF-VOC-aug \cite{liu2016richer}                                      &         .806           &                .823       &                    $30^{\dagger}$\\
 RCF-MS-VOC-aug \cite{liu2016richer}                              &         .811           &                .830       &                     $10^{\dagger}$ \\
 CED \cite{wang2017deep}                               &          .794          &                 .811       &                    $30^{\dagger}$\\
 CED-MS \cite{wang2017deep}                       &          .803          &                 .820       &                    $10^{\dagger}$\\
 CED-MS-VOC-aug \cite{wang2017deep}           &          \textbf{.815}       &                 .833        &                   $10^{\dagger}$\\
             
\hline
\textbf{Ours}                                                           &          .800           &                .816       &                    $30^{\dagger}$\\
\textbf{Ours-VOC-aug}                                        &         .808           &                 .824     &                      $30^{\dagger}$\\
\textbf{Ours-MS-VOC-aug}                                &         \textbf{.815}           &               \textbf{ .834 }    &                      $10^{\dagger}$\\
\hline
\end{tabular}
\label{bsds numerical comparison}
\end{table}
We first look at the qualitative result in Figure \ref{figs of comparison on BSDS}. 
RCF and CED are the leading edge detectors at present. 
Especially, CED shares the same aim with our method, which is to solve the issue of boundary crispness. 
Comparing to the other methods, our approach shows a clear advantage in quality of edge maps which are cleaner and sharper. 
Consider the `cow' in the third example. Our method is able to precisely match its contour, 
whereas RCF and CED incur much more blurry and noisy edges. 
The qualitative comparisons suggest that our method generates sharper boundaries.

The quantitative results are summarized in Table \ref{bsds numerical comparison}. 
Figure \ref{bsds_curve} shows Precision-Recall curves of all methods. 
Note that, all the results have been post processed (using NMS) before evaluation. 
Without extra training data and the multi-scale testing, our method already outperforms most of the state-of-the-art edge detectors. 
By means of extra training data, our single-scale model achieves a significant improvement on ODS f-score from .800 to .808. 
With the multi-scale testing, our method achieves the same top performance with CED. 
However, CED adopted both the train and validation set for training while we only use the train set. 

In addition to this, we evaluate the non-NMS results of CED (single-scale, without extra training data) 
and obtain  the performance of ODS f-score of .655, OIS f-score of .662. 
The result is far behind our single-scale non-NMS performance (ODS f-score of .693). 
Another advantage of our method is that our detector is able to run in real time. The single scale detector can operate at 30FPS on a GTX980 GPU. 
Since our method is simple, effective and very fast, it is easy to be used along with  high-level vision tasks such as image segmentation.
\begin{figure*}[t]
\centering
\begin{minipage}{0.22\textwidth}
{\includegraphics[width=\linewidth,height=1.2cm]{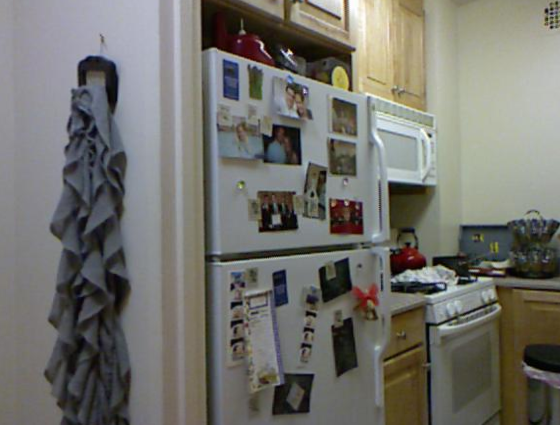}}
\end{minipage}
\begin{minipage}{0.22\textwidth}
{\includegraphics[width=\linewidth,height=1.2cm]{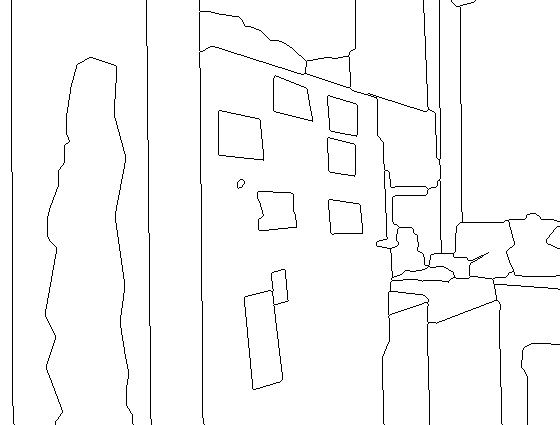}}
\end{minipage}
\begin{minipage}{0.22\textwidth}
{\includegraphics[width=\linewidth,height=1.2cm]{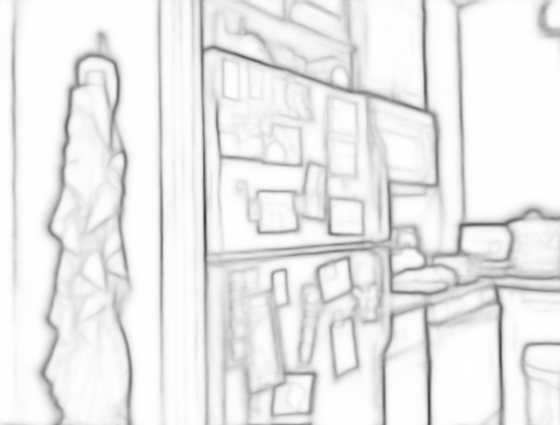}}
\end{minipage}
\begin{minipage}{0.22\textwidth}
{\includegraphics[width=\linewidth,height=1.2cm]{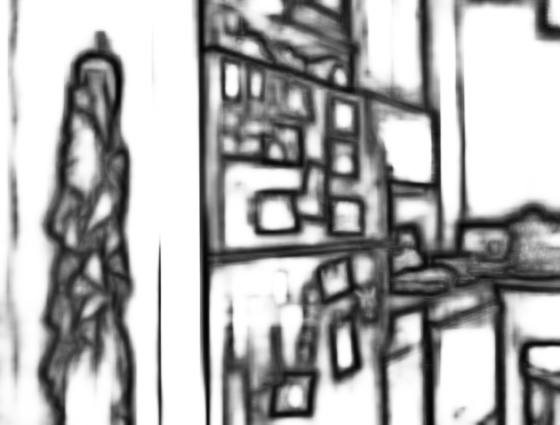}}
\end{minipage}
\vskip 0.05in
\begin{minipage}{0.22\textwidth}
{\includegraphics[width=\linewidth,height=1.2cm]{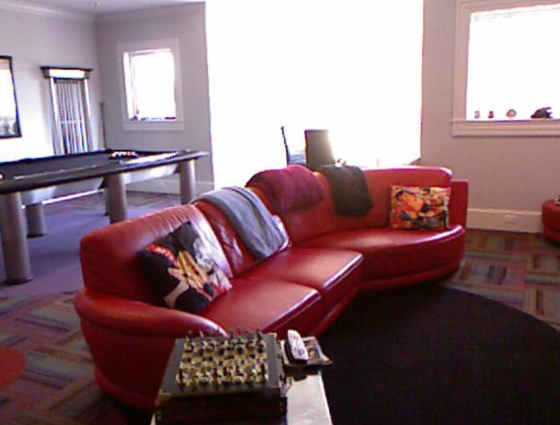}}
\end{minipage}
\begin{minipage}{0.22\textwidth}
{\includegraphics[width=\linewidth,height=1.2cm]{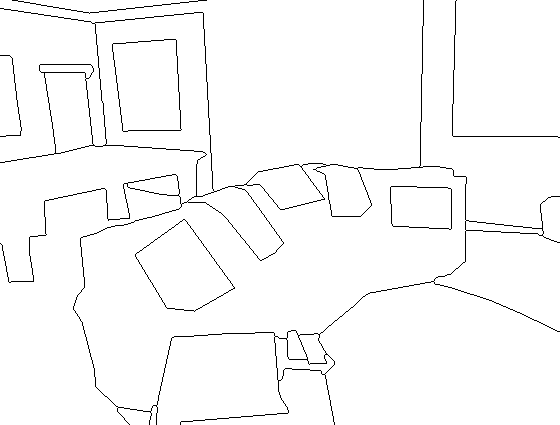}}
\end{minipage}
\begin{minipage}{0.22\textwidth}
{\includegraphics[width=\linewidth,height=1.2cm]{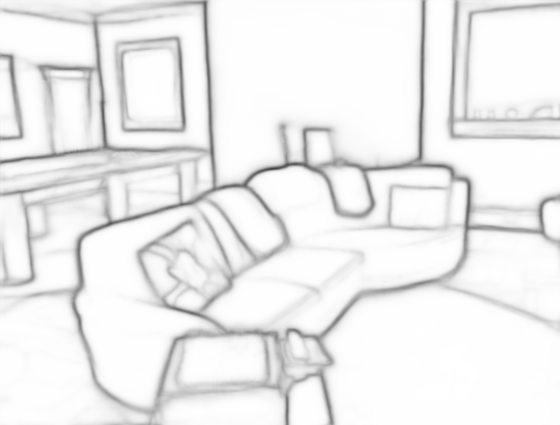}}
\end{minipage}
\begin{minipage}{0.22\textwidth}
{\includegraphics[width=\linewidth,height=1.2cm]{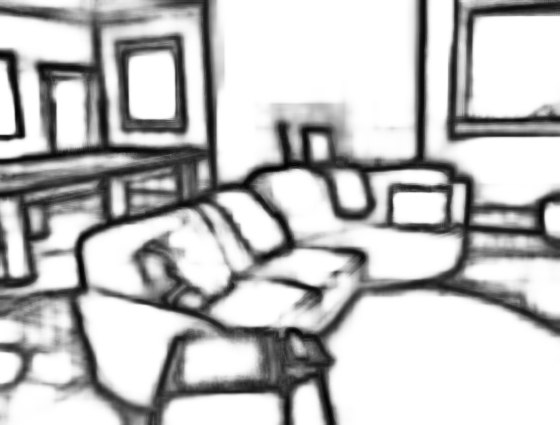}}
\end{minipage}
\vskip 0.05in
\begin{subfigure}{0.22\textwidth}
{\label{main:a2}\includegraphics[width=\linewidth,height=1.2cm]{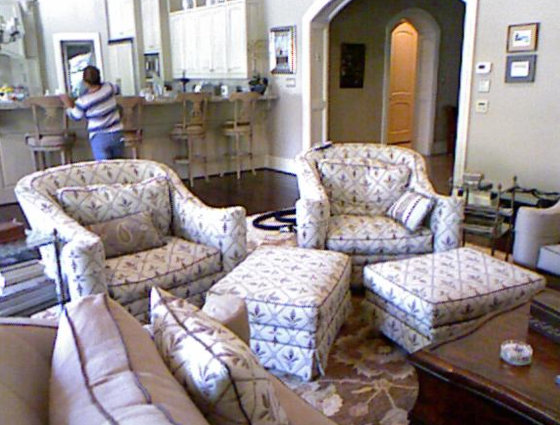}}
\caption{Input Image}
\end{subfigure}
\begin{subfigure}{0.22\textwidth}
{\label{main:e2}\includegraphics[width=\linewidth,height=1.2cm]{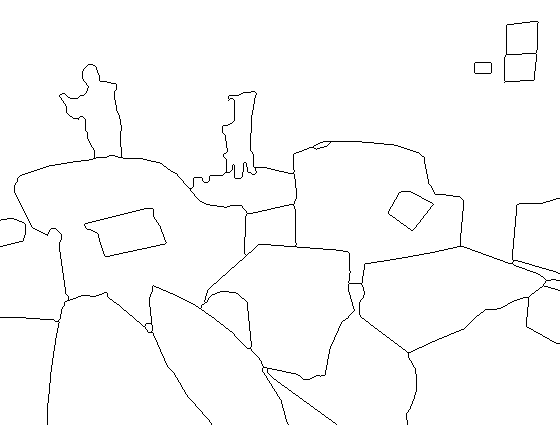}}
\caption{GT}
\end{subfigure}
\begin{subfigure}{0.22\textwidth}
{\label{main:b2}\includegraphics[width=\linewidth,height=1.2cm]{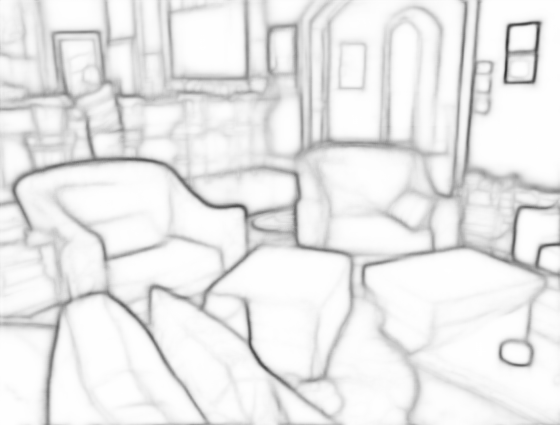}}
\caption{Ours}
\end{subfigure}
\begin{subfigure}{0.22\textwidth}
{\label{main:d2}\includegraphics[width=\linewidth,height=1.2cm]{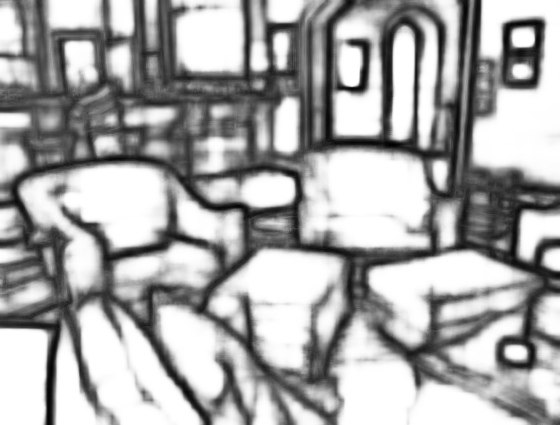}}
\caption{RCF \cite{liu2016richer}}
\end{subfigure}
\caption{State-of-the-art comparisons on NYUDv2. From left to right:
 (a) is the original image, (b) is the groundtruth, (c) is the prediction of the proposed method, (d) is the result of the RCF detector. 
 Note that the predictions of RCF and the proposed method are trained only on the RGB data. No postprocessing is applied.}
\label{figs of comparison on NYU}
\end{figure*}
\begin{table}[t]
\renewcommand{\arraystretch}{1.05}
\caption{Results on the NYUDv2 dataset. $\dagger$ means GPU time.}
\centering
\begin{tabular}{ c|c|c|c  }
 \hline
 Method                                                             &                  ODS          &              OIS      &                FPS \\
 \hline
OEF \cite{hallman2015oriented}  &                  .651            &             .667      &                1/2\\
 gPb-UCM \cite{arbelaez2011contour}  &                   .631           &              .661      &                1/360\\
 gPb+NG \cite{gupta2013perceptual}    &                   .687            &             .716       &                1/375\\
 SE  \cite{dollar2015fast}                            &                    .695           &                .708     &                5   \\
 SE+NG+ \cite{gupta2014learning}        &                    .706           &               .734      &                1/15\\
 \hline
 HED-RGB \cite{xie2015holistically}     &                    .720           &               .734      &                20$\dagger$\\
 HED-HHA \cite{xie2015holistically}     &                    .682           &               .695      &                20$\dagger$\\
 HED-RGB-HHA \cite{xie2015holistically}     &                    .746           &               .761      &                10$\dagger$\\
 \hline
 RCF-RGB \cite{liu2016richer}                                      &         .729           &                .742       &                    $20^{\dagger}$\\
 RCF-HHA \cite{liu2016richer}                              &         .705           &                .715       &                     $20^{\dagger}$ \\
 RCF-RGB-HHA \cite{liu2016richer}                              &         .757           &                .771       &                     $10^{\dagger}$ \\            
\hline
Ours-RGB                                                                                &         .739         &                 .754     &                      $30^{\dagger}$\\
Ours-HHA                                                                           &              .707           &                .719    &                      $30^{\dagger}$\\
Ours-RGB-HHA                                                                &             \textbf{ .762}          &      \textbf{.778}    &                      $15^{\dagger}$\\
\hline
\end{tabular}
\label{nyud numerical comparison}
\end{table}
\subsection{NYUDv2 dataset}
The NYU depth dataset \cite{Silberman:ECCV12} is a large depth benchmark for indoor scenes, 
which is collected by a Microsoft Kinect sensor. 
It has a densely labeled dataset (every pixel has a depth annotation) which has 1449 pairs of aligned RGB and depth images. 
Gupta \etal \cite{gupta2013perceptual} processed the data to generate edge annotation and split the dataset into 381 training images, 414 validation images, and 654 testing images. 
We follow their data-split setting and change several hyper-parameters of our method for training: mini-batch size (26), image resolution ($480\times480$). The maximum 
tolerance allowed for correct matches of edge prediction in evaluation is increased from .0075 to .011, as used in \cite{xie2017holistically,liu2016richer,dollar2015fast}. 
We compare against the state-of-the-art methods which include 
OEF \cite{hallman2015oriented}, gPb-UCM \cite{arbelaez2011contour}, gPb+NG \cite{gupta2013perceptual}, SE \cite{dollar2015fast}, 
SE+NG+ \cite{gupta2014learning}, HED \cite{xie2015holistically} and RCF \cite{liu2016richer}. 

Motivated by the previous works \cite{xie2015holistically,liu2016richer}, we leverage the depth information to improve performance. 
We employ the HHA feature \cite{gupta2014learning} in which 
the depth information is encoded into three channels: horizontal disparity, height above ground, and angle with gravity. 
The way of employing the HHA feature is straightforward. 
We simply train two versions of the proposed  network, one on the RGB data, another on HHA feature images. 
The final prediction is generated by directly averaging the output of the RGB model and HHA model.
\begin{figure}
    \centering
    \begin{minipage}{0.45\textwidth}
        \hspace*{-1cm}
        \vspace*{0.9cm}
        \includegraphics[scale=0.5]{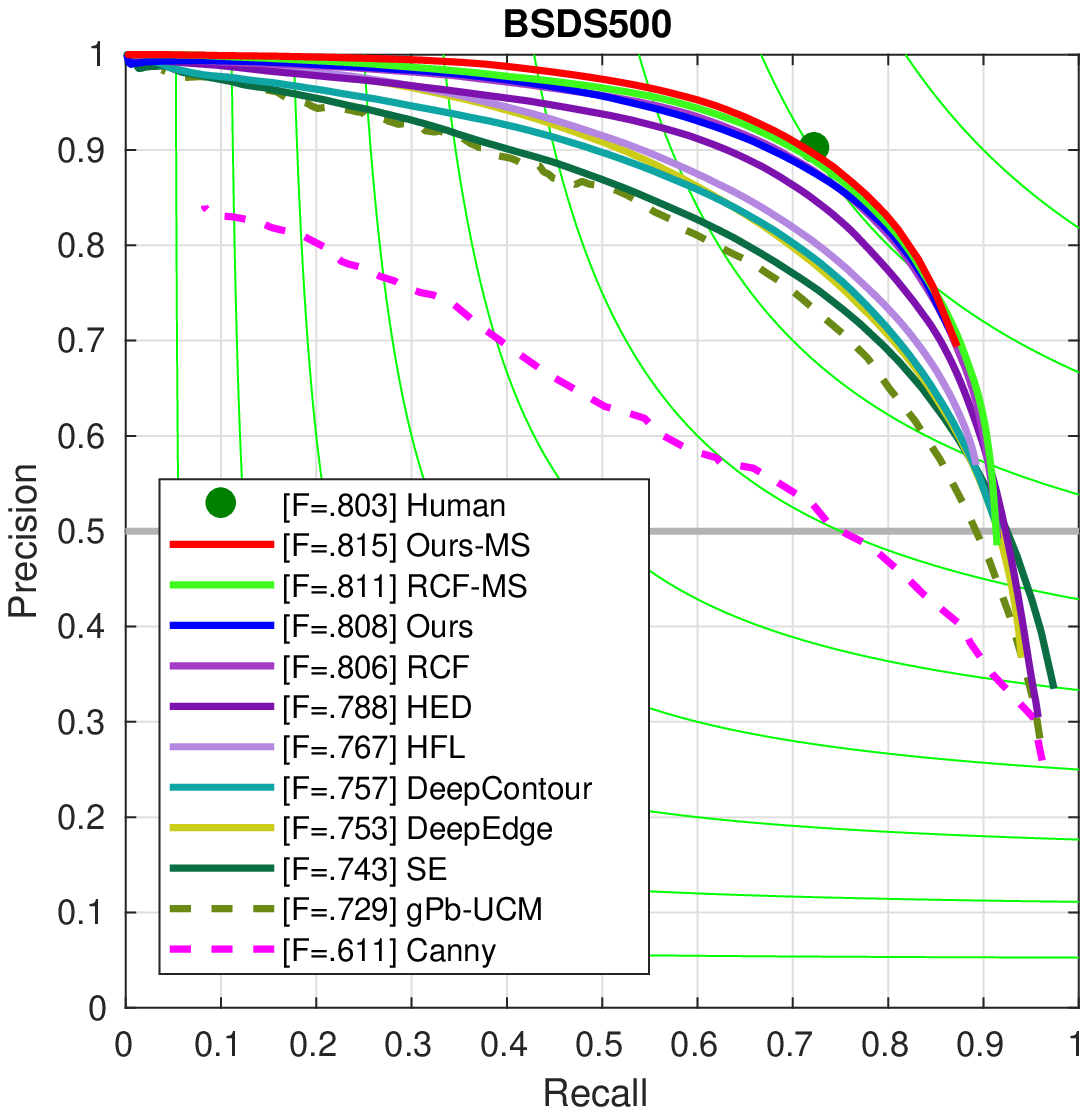} %
        \caption{Precision/recall curves on the BSDS500 dataset. Our method ahieves the best result (ODS=.815).}
       \label{bsds_curve}
    \end{minipage}\hfill
    \begin{minipage}{0.45\textwidth}
       \hspace*{-1.1cm}
        \includegraphics[scale=0.5]{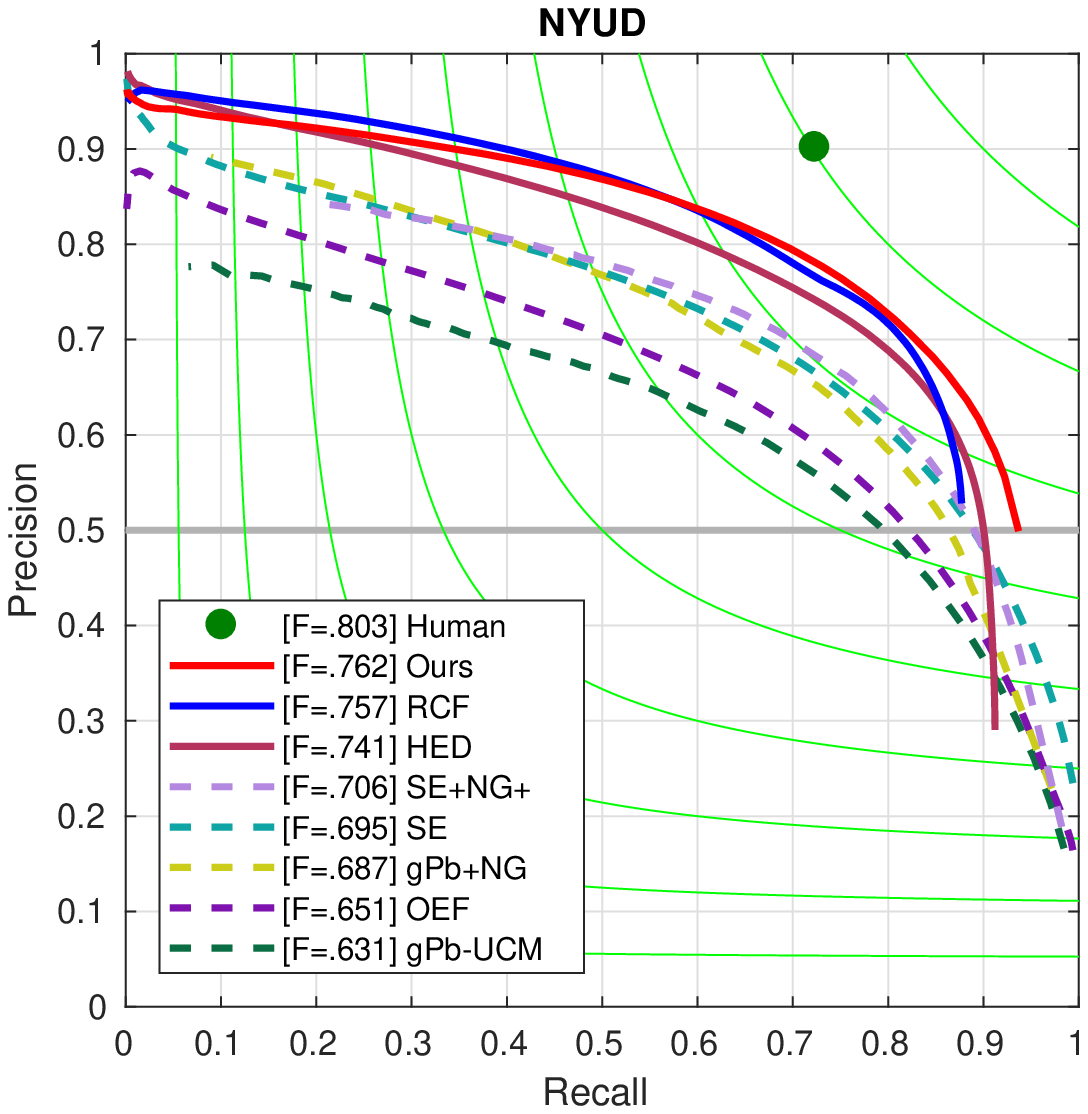} %
        \caption{Precision/recall curves on the NYUDv2 dataset. Our method trained with the RGB data and the HHA feature achieves the best result (ODS=.762).}
        \label{nyud_curve}
    \end{minipage}
\end{figure}
We show the quantitative results in Table \ref{nyud numerical comparison} and 
the precision-recall curve in Figure \ref{nyud_curve}. Our method achieves the best performance of ODS F-score $.762$. 
The qualitative results in Figure \ref{figs of comparison on NYU} show consistent performance with those of the experiments on BSDS 500. 
Our prediction produces sharper boundaries against the leading competitor RCF, 
which demonstrates the effectiveness of our method.
\section{Conclusions}
In this work, we have presented a simple yet effective method for edge detection which achieves state-of-the-art results. 
We have shown that it is possible to achieve excellent boundary detection results using a carefully designed loss function 
and a simple  convolutional encoder-decoder network. 
 
In future work, we plan to extend the use of the edge detector to the tasks like object detection and optical flow 
which have the requirement of boundary sharpness and a fast  processing  speed.

\bibliographystyle{splncs}
\bibliography{egbib}
\end{document}